\documentclass[runningheads]{llncs}
\usepackage{graphicx}
\usepackage{color}
\usepackage{amssymb}
\begin{document}
\title{The Impact of Activation Sparsity on Overfitting in Convolutional Neural Networks}
%
\titlerunning{Neural Activation Sparsity}
%
\author{Karim Huesmann \and
Luis Garcia Rodriguez \and
Lars Linsen \and
Benjamin Risse$^\ast$
}
\authorrunning{Huesmann et al.}
%
\institute{Faculty of Mathematics and Computer Science, University of Muenster, Germany 
\email{\{karim.huesmann, luis.garcia, linsen, b.risse\}@uni-muenster.de}\\
$^\ast$ Corresponding Author (\url{https://cvmls.uni-muenster.de})}
\maketitle              
\begin{abstract}
Overfitting is one of the fundamental challenges when training convolutional neural networks and is usually identified by a diverging training and test loss.
The underlying dynamics of how the flow of activations induce overfitting is however poorly understood. 
In this study we introduce a perplexity-based sparsity definition to derive and visualise layer-wise activation measures.
These novel explainable AI strategies reveal a surprising relationship between activation sparsity and overfitting, namely an increase in sparsity in the feature extraction layers shortly before the test loss starts rising.
This tendency is preserved across network architectures and reguralisation strategies so that our measures can be used as a reliable indicator for overfitting while decoupling the network's generalisation capabilities from its loss-based definition.
Moreover, our differentiable sparsity formulation can be used to explicitly penalise the emergence of sparsity during training so that the impact of reduced sparsity on overfitting can be studied in real-time.
Applying this penalty and analysing activation sparsity for well known regularisers and in common network architectures supports the hypothesis that reduced activation sparsity can effectively improve the generalisation and classification performance.  
In line with other recent work on this topic, our methods reveal novel insights into the contradicting concepts of activation sparsity and network capacity by demonstrating that dense activations can enable discriminative feature learning while efficiently exploiting the capacity of deep models without suffering from overfitting, even when trained excessively.

\end{abstract}
\section{Introduction}
\label{sec:intro}
In recent years deep convolutional neural networks (CNNs) achieved state-of-the-art performances in most computer vision applications~\cite{khan2019survey,tu2019survey}. 
The ultimate goal of training neural networks networks is to achieve high performance measures while avoiding the generalisation error, which is defined by the difference between the training and test set loss~\cite{zhang2017understanding}.
This error estimate requires that the trained model is independent of the test set (independence hypothesis)~\cite{werpachowski2019detecting}.
If the independence hypothesis holds~\cite{werpachowski2019detecting}  and generalisation error is high, then the model is suspected to have an inappropriately high variance and is therefore overfitted to the training data~\cite{zhang2018study}.
In fact, overfitting appears to be one of the fundamental challenges in training deep CNNs since models often tend to learn too specific features of the training set~\cite{gavrilov2018preventing,ayinde2019correlation}.

\subsection{Related Work}
In the past, a variety of strategies have been proposed to prevent overfitting, which can be roughly categorised into: (i) increasing the amount of training data; (ii) reducing the models' capacity; (iii) regularising the model parameters; and (iv) early stopping procedures.
Increasing the training set can either be done by collecting additional data or by augmenting the existing set~\cite{cubuk2019autoaugment,guo2019mixup}.
The reduction of the capacity is usually done by explicitly pruning learnable parameters~\cite{goodfellow2016deep,liu2017learning,klemm2019deploying,molchanov2016pruning}.
In a similar fashion, regularisation can also be used to decrease the capacity of the model~\cite{nowlan1992simplifying,Ioffe2015}.
For example~\cite{nowlan1992simplifying} and~\cite{ayinde2017deep}
both aim to extenuate the model complexity by using weight decay and it has also been shown that classical L1 regularisation tends to less complex models over time~\cite{yaguchi2018adam,mehta2019implicit}.
In contrast, decorrelation-based regularisers aim to employ the given capacity by reducing redundancies based on hidden features~\cite{bengio2009slow,ayinde2019regularizing} or activations~\cite{bao2013incoherent,cogswell2015reducing}, whereas entropy-based regularisation uses the output distribution of the network to reduce overfitting~\cite{miller1996global,pereyra2017regularizing}.
Alternative regularisation strategies incorporate additional layers.
The two most common examples are dropout~\cite{Srivastava2014} and batch normalisation~\cite{Ioffe2015}.
Finally, early stopping has been studied for more than a decade to end the training before the generalisation error gets too high~\cite{prechelt1998early} 
and has recently been studied in the presence of label noise~\cite{li2019gradient}.

An important concept often related to overfitting is neural network sparsity, which is interpreted as an indicator of success in learning discriminative features~\cite{ahmad2019can} while filtering out irrelevant information, in analogy to sparsity observed in the neocortex~\cite{vinje2000sparse}.
Consequently, many of the previously mentioned pruning and regularisation techniques assume that sparse models suffer less from generalisation error~\cite{goodfellow2016deep,liu2017learning,klemm2019deploying,molchanov2016pruning,nowlan1992simplifying,ayinde2017deep,yaguchi2018adam,mehta2019implicit}.
In addition, several empirical investigations have been done to demonstrate beneficial effects of sparsity on the performance of deep neural networks~\cite{changpinyo2017power,gale2019state,frankle2018lottery}. 
However, recent experiments suggest that our comprehension of sparsity is insufficient to fully understand its underlying relationship to generalisation errors~\cite{liu2018rethinking,zhou2019deconstructing}.
Interestingly, even though sparsity inevitably induces an underutilisation of the network's capacity~\cite{ayinde2019regularizing} it has never been considered to be used to explain overfitting.

\subsection{Contribution} \label{sec:contribution}
Usually sparsity is referred to the property of zero-valued weights~\cite{gale2019state} and is distinguished from \textit{activation sparsity} which counts the number of zeros after applying a non-linearity (in general ReLU)~\cite{10.1109/hpca.2018.00017,yang.2019}.
Most existing work on (activation) sparsity focuses on its beneficiary effects such as improved inference performances or the robustness to adversarial attacks and noise~\cite{ahmad2019can,guo2018sparse,frankle2018lottery}.
Here we propose a novel activation sparsity definition which reveals a surprising relationship between sparsity and overfitting.
Our analysis suggests that high activation sparsity is a reliable indicator for an underutilisation of the potential network capacity and therefore limits the generalisation capabilities of the model which induces overfitting.
To study this hypothesis we propose:
\begin{enumerate}
  \item a perplexity-based activation sparsity definition which uses network features across neural receptive fields, yielding a novel layer-wise (i.e. targeted) sparsity measures, called \textit{neural activation sparsity} (NAS);
  \item targeted explainable AI (XAI) visualisation strategies to identify neural activation sparsity in both, convolutional and fully connected layers in real-time;
  \item a regularisation strategy derived from our neural activation sparsity definition which can be used to penalise or reward neural activation sparsity in a targeted manner. 
\end{enumerate}
Using these techniques, we demonstrate that an increase in neural activation sparsity in all layers except the prediction layer indeed coincides with test loss-based overfitting measures.
Therefore, our XAI measures can be used as an indicator for overfitting, which is independent of loss values, thus enables the detection of generalisation errors even if the independence hypothesis of the training and test data is violated.
To further demonstrate this relationship, we counteract neural activation sparsity in a layer-specific fashion using a novel regularisation strategy:
by penalising neural activation sparsity in the feature extraction layers we can prevent the network from overfitting while increasing the overall accuracy.
We elaborate our findings by performing several experiments including common network architectures (including VGG16, ResNet50 and Xception) and regularisation strategies (L1, L2, dropout and batch normalisation).
Interestingly, targeted neural activation sparsity regularisation prevents overfitting even in low training data / high model size scenarios and outperforms conventional regularisation techniques. 
In line with other recent work on activation sparsity~\cite{liu2018rethinking,zhou2019deconstructing} we demonstrate that neural network sparsity must not be a desirable property by definition. 
Instead salient and discriminative feature learning while exploiting the full capacity of the model appears to be another important dimension to effectively train deep convolutional neural networks. 

\section{Methods}
\label{sec:methods}

\subsection{Activation Sparsity Definition}
\begin{figure}[t]
  \centering 
  \includegraphics[width=\linewidth]{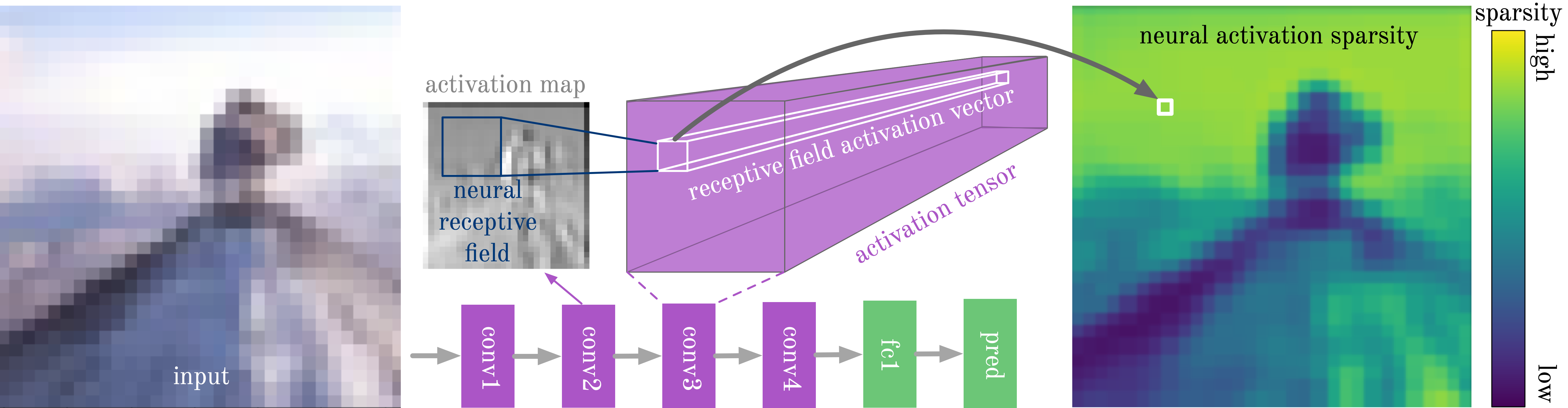}
  \caption{NAS and NAS Heatmap generation.
  Left: sample input image.
  Middle: schematic network architecture and sketch of an exemplary proximate receptive field from the conv2 layer and the corresponding receptive field activation vector  in conv3 layer. 
  Right: NAS Heatmap extracted by calculating the activation sparsity on the depicted  receptive field activation vector.
  }
  \label{fig:sparsity_visualization}
\end{figure}

Neural network sparsity is usually defined by the amount of weights which are exactly or close to zero~\cite{gale2019state}.
Even though this quantification is inspired by observations made in real brains~\cite{vinje2000sparse}, neuroscientific sparsity definitions are based on neuronal activation patterns~\cite{seelig2010two}.
In a similar fashion sparse information propagation in the forward path of artificial neural networks can be the result of zero-valued weights or zero-valued input coefficients.
Motivated by this observation we introduce a novel sparsity indicator that is purely based on the feature map activations which we refer to as \textit{neural activation sparsity} (NAS).
We note that our sparsity definition differs from existing activation sparsity definitions in two ways. 
Firstly, conventional activation sparsity measures count the number of zeros after the use of the non-linearity~\cite{10.1109/hpca.2018.00017,yang.2019}, whereas our metric is calculated before the nonlinearity is applied.
As a consequence NAS is independent of the activation function (i.e. less biased) and captures both, weight- and data-induced sparsifications (i.e. more sensitive).
Secondly, we measure relative neural activity values in comparison to all other feature map activations at a given neural receptive field. 
The neural receptive field is defined as the region of a layer’s direct input that a filter is being affected by (see Figure~\ref{fig:sparsity_visualization}).

More formally, a layer with $D$ filters creates a $D \times M \times N$ shaped feature tensor. 
Thus, the input of this layer consists of $M \cdot N = R$ receptive fields. 
Let $x_{d',i,j,m,n}$ be a pixel at position $(i, j)$ and channel $d'$ of receptive field $r_{m,n}$ .
The corresponding weight of filter $f_d$ which affects this pixel is given as $w_{d',i,j,d}$.
The number of pixels in $r_{m,n}$ is equal to the number of weights in $f_d$.
Hence, the linear activation $a_{m,n,d}$ created by $f_d$ when applied to $r_{m,n}$ is defined as
\begin{equation}
a_{m,n,d} = \sum_{d',i,j} w_{d',i,j,d} \cdot x_{d',i,j,m,n} \ .
\end{equation}

To improve readability, we denote the results of this linear filtering as receptive field activation vectors $\textbf{a}_k \in \mathbb{R}^D, k \in [0,R-1]$, where $k$ is a linear index over all receptive fields computed by $k = m \cdot N + n$.

In order to define activation sparsity based on $\mathbf{a}_k$ we first transform its components into a probability-like distribution by using the Softmax function:
\begin{equation}
p_l(\textbf{a}_k) = \frac{e^{a_k^l}}{\sum_m e^{a_k^m}} \ , \ l \in [1, ..., D] 
\end{equation} 
where $a_k^l$ is the $l$-th component of $\textbf{a}_k$. 
Subsequently, we calculate the perplexity for each normalised neural receptive field by 
\begin{equation}
\label{equ:perplexity}
    \rho_k = e^{H_k}
\end{equation}
with
\begin{equation}
\label{eq:entropy}
H_k = H(\textbf{a}_k) = - \sum_{l=1}^D p_l(\textbf{a}_k) \ ln(p_l(\textbf{a}_k)) \ .
\end{equation}
By definition, perplexity results in values in a range between $[1, D]$.
Next, we define a perplexity score $\tau$ by rescaling $\rho$ into values $\le 1$: 
\begin{equation}
    \tau_k = \frac{\rho(\textbf{p})_k}{D} ,
\end{equation}
where $\tau_k \in [\frac{1}{D}, 1]$. Based on this definition, the smaller/larger $\tau_k$ is, the less/more filters are activated at the neural receptive field $k$. Finally, we can define the neural activation sparsity by
\begin{equation}
    \label{equ:sparsity}
    s_k = 1-\tau_k \ .
\end{equation}
This activation sparsity measure reaches its maximum value $s_k = 1 - \frac{1}{D}$, 
if only one of the given $D$ feature maps returns an activation $a_l$, $ a_l \in \textbf{a}_k$, $p_l(\textbf{a}_k) \neq 0$, indicating lowest possible  activation density for a given neural receptive field.
The minimum value $s_k = 0$ can be reached, if all $D$ feature maps return the same activation $a_l = a_m, l \neq m$, which is equivalent to the highest possible activation density and thus minimum activation sparsity.

\subsection{Activation Sparsity Visualisation}
\begin{figure}[t]
   \centering 
   \includegraphics[width=1.\linewidth]{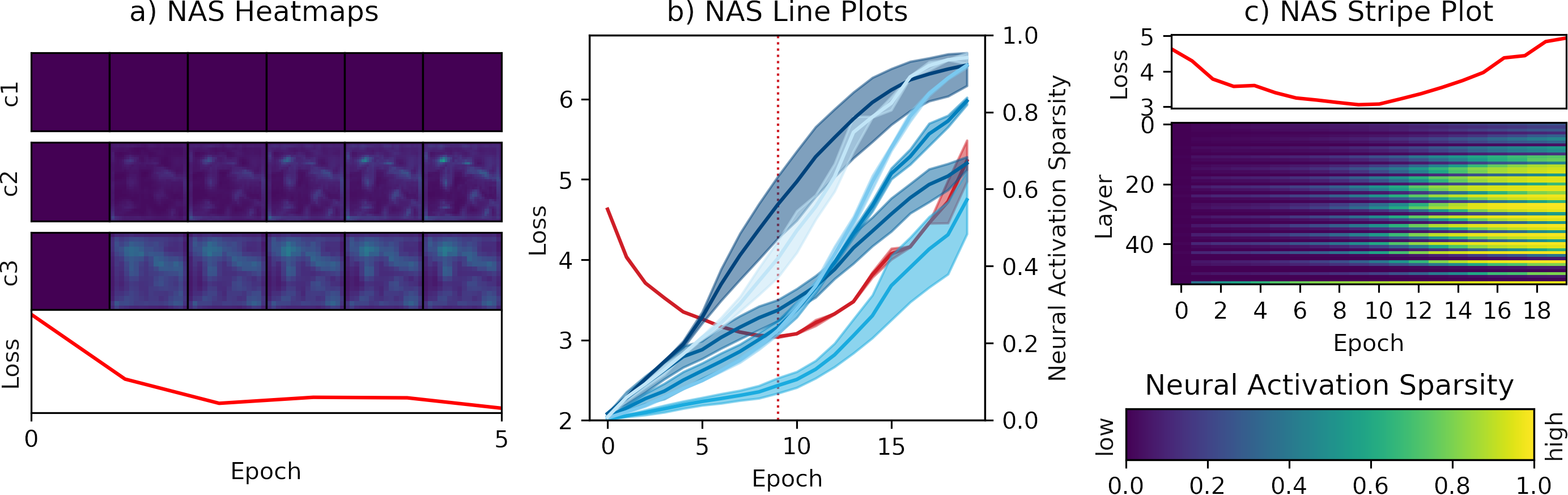}
   \caption{Visualisations of layer NAS over time at three levels of aggregation with corresponding test loss (red): (a) \emph{NAS Heatmaps} showing spatial sparsity distributions for 3 layers. 
   (b) \emph{NAS Line Plot} showing 
   blue lines (and bands) of layer NAS values. 
   (c) \emph{NAS Stripe Plot} for 50 layers encoding layer NAS in 1D colour coded stripes.
    }
   \label{fig:vis}
\end{figure}
In order to enable a temporal visualisation of layer NAS for a variety of CNN architectures we propose three different approaches that visually represent the time series at different levels of aggregation using geometric (lines) and colour (heatmap) encodings. 
These strategies are a trade-off between the level of visualised details and the underlying network depth.

The most detailed visualisation can be achieved by spatially localised colour-encodings of each NAS value for all layers.
By applying the newly defined activation sparsity, we can calculate $s_k$ for each position $k$ in all layers, indicating how many of the layer's filters produce a comparatively high activation for the given filter input.
Encoding and re-ordering all $s_k$ into a 2D heatmap we can visualise the activation sparsity in a very detailed manner for a given input as illustrated in Figure~\ref{fig:vis} (a):
High sparsity measures are given in yellow indicating that only a few filters produced high activations at the location of the respective neural receptive field.
Likewise, the dark pixels indicate many highly activated filters (i.e. low NAS measures) which naturally occurs at contrast-rich locations such as edges and textured regions.
Note that the same visualisation strategy can be used for fully connected layers, with the only difference that the resulting heatmap consists of a single value.

Even though NAS heatmaps enable spatial reasoning within the activation maps, this visualisation can become overloaded given many layers or training epochs.
In order to provide a more concise overview of NAS values for all layers we therefore introduce the NAS Line Plot which visualise the median NAS values over all neural receptive fields for a given layer into the same illustration as shown in Figure~\ref{fig:vis} (b).
Each line therefore represents the change in activation sparsity (y-axis) of an individual layer over time (x-axis), where brightness of the colour encodes the depth of the layers (early layers in dark blue and deep layers in bright blue). Furthermore, we are able to visualise uncertainties between multiple trainings by including a band, which in fact makes the plot a functional box plot over time (minimum, median, maximum).
NAS Line Plots allow for effectively reading off the NAS values for each layer at each time step including their uncertainty, but they are limited to a certain amount of layers since very deep networks will lead to occluded curves.

Therefore we also provide a visualisation technique which further abstracts from the measured NAS values by means of a colour-coded  1D NAS heatmap over time (x-axis), which we refer to as NAS Stripe Plot. 
The NAS Stripe Plots are sorted by increasing layer depth (y-axis).
As can be seen in Figure~\ref{fig:vis}(c) this visualisation encodes the least details but can be used to illustrate NAS dynamics over dozens of layers.

In Figure~\ref{fig:vis} all NAS visualisations are superimposed with the respective test loss over time (red line) to demonstrate the impact of activation sparsity on overfitting. For NAS Heatmaps and Stripe Plots, we always applied the same colour map as shown in Figure~\ref{fig:vis}, encoding NAS values from low (dark blue, $s_k = 0$) to high (yellow, $s_k \approx 1$). 
Moreover these visualisations can be calculated dynamically during training so that these XAI strategies can offer deep insights into activation sparsity of a given network in real-time.

\subsection{Activation Sparsity Regularisation}
\label{sec:regulariser}
To test our hypothesis, that NAS is related to overfitting, we introduce a penalty term that prevents the network layers from producing sparse activations. 
This term is directly derived from our aforementioned activation sparsity definition and utilises the fact that perplexity is differentiable in our given ranges so that it can be used as an activity regulariser.
More formally, our regularisation penalty term $\mathcal{L}_{s}$ is defined by
\begin{equation}
\label{eq:sparsity-reg}
\mathcal{L}_{s} = - \sum_{i} \; \lambda_i \sum_{k=0}^{r_i}  \rho_{k}^{i} ,
\end{equation}
where $i$ loops through all layers of the network, $r_i$ is the number of receptive fields in the respective layer and $\rho_{k}^{i}$ refers to $\rho_{k}$ as defined in Equation~\ref{equ:perplexity}. 
In the following, this regularisation is referred to as \textit{NASReg}.

As perplexity reaches its maximum when all filters are activated in the same way, this regulariser can have the tendency to produce highly correlated filters. 
The trivial solution for $\mathcal{L}_{s}$ to be minimised is therefore to generate identical filter responses.
As identical filters reduce the predictive power of neural networks, this effect has to be counterbalanced by preventing high filter correlations. 
Therefore we have to set $\lambda_i \geq 0$ in a way, that the perplexity is not reaching it's theoretical maximum.

\subsection{Experimental Design}
\label{sec:design}
In our first set of experiments we used an intentionally simple base line architecture called \textit{VanillaNet}. 
This basic network comprises two conv-conv-pool blocks followed by two fully connected layers and the layer sizes of conv1, conv2, conv3, conv4, and fc1 are 256, 256, 512, 512, 1024 respectively.
ReLU is used as activation function across all layers except the last fully connected layer. 

Next, we analysed our XAI strategies in the context of different regularisation techniques.
In particular, we extended the VanillaNet by including L1, L2, dropout and batch normalisation (called \textit{L1Net, L2Net, DropNet} and \textit{NormNet} respectively) since these techniques are known to have an impact on sparsity.
For DropNet we set the dropout rate to 0.3 for layers conv1 to fc1 and 0.5 between fc1 and prediction layer. 

To study the generalisability of our metrics and visualisation strategies for state-of-the-art architectures we evaluated three very deep CNNs, namely VGG16 \cite{simonyan2014very}, ResNet50~\cite{he2016deep} and Xception~\cite{chollet2017xception}.
Moreover, the behaviour of ResNet with and without batch normalisation is analysed to further evaluate the impact of regularisation on activation sparsity of deep nets.

For a better interpretability and to reduce side effects of the experiments we have chosen a non-adaptive optimiser. 
Accordingly, we used vanilla Stochastic Gradient Decent without momentum and a fixed learning rate of $0.01$ without learning rate scheduling.
This allows us to carry out a uniform training over the entire course of the experiments.
For the classification we are optimising the categorical cross-entropy loss. 
The experiments are trained up to $300$ epochs with a batch size of $32$.

During the training, we calculate NAS for each convolution and fully connected layer individually. 
For most of our experiments we chose the cifar-100 dataset ($50,000$ train and $10,000$ test images), because it is very susceptible to overfitting.
For one particular experiment we intentionally violated the independence hypothesis of our test dataset by creating a new testing set consisting of $5,000$ train and $5,000$ test images, which we mixed to generate a distorted test loss.
We conducted a hyperparameter search to find appropriate values for $\lambda_i$ which resulted in a strong correlation between $\lambda_i$ and the size of the layers.
Therefore we reduced the mean NAS via $\lambda_i=\frac{1}{r_i}$ for each layer $i$ in all experiments.
In order to add uncertainty to the results, we repeat each training 3 times.

\section{Results}
\label{sec:results}

\subsection{Relationship Between Overfitting and Activation Sparsity}
\label{sec:results-relationship}
\begin{figure}[t]
   \centering 
   \includegraphics[width=1.0\linewidth]{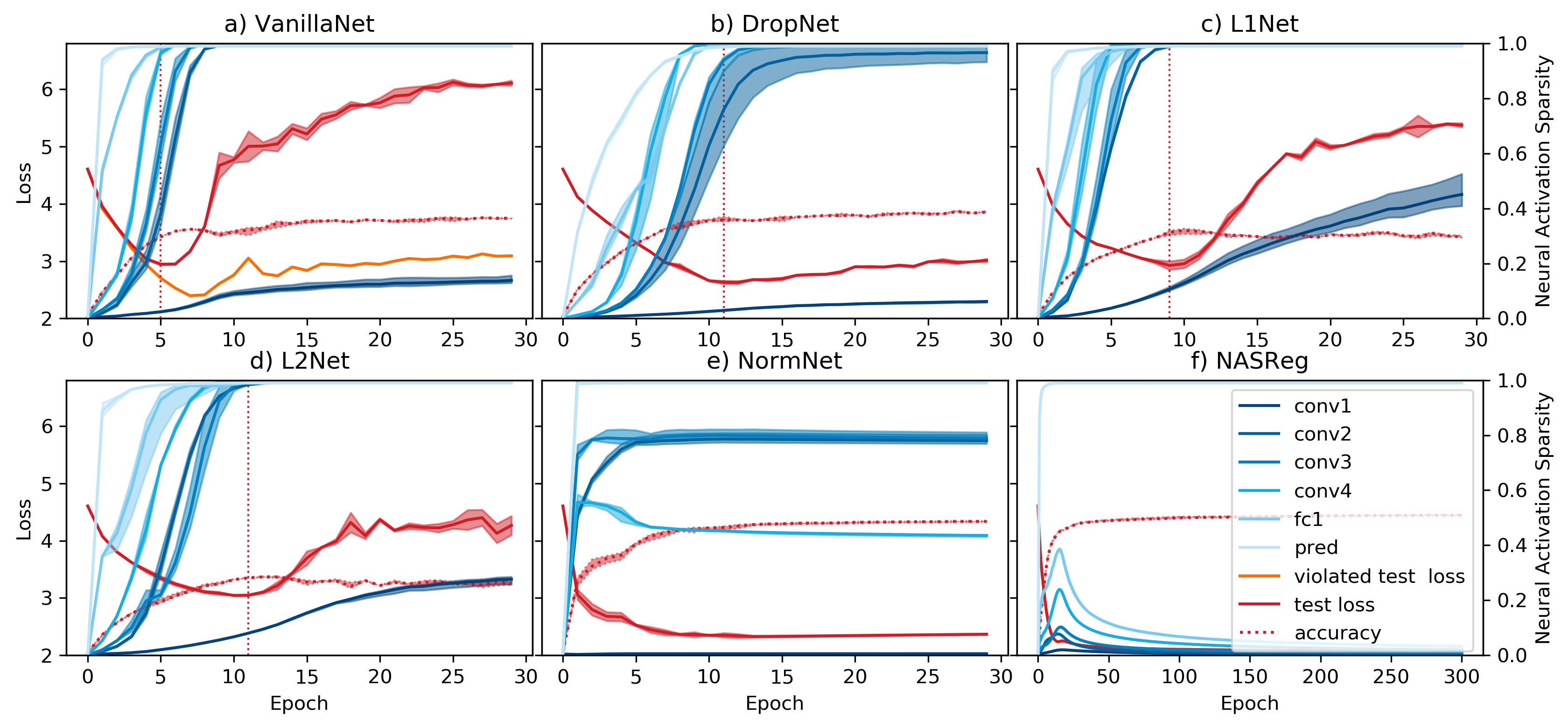}
   \caption{NAS Line Plots for Test loss (left axis) and median NAS (right axis) over time.
   The dashed red lines indicate the moment of overfitting derived from the test loss. All trainings are carried out on cifar-100 dataset.
  (a) - (e) networks are trained for 30 epochs whereas the NASReg network in (f) is trained for 300 epochs. 
   }
   \label{fig:fig4}
\end{figure}

To test if classical test loss based generalisation errors coincide with our NAS definition, we analysed the NAS Line Plots for multiple experiments.
In Figure~\ref{fig:fig4}(a) the resultant NAS Line Plots of VanillaNet on cifar-100 can be seen, which are not affected by any regularisation, thus allowing a straight-forward interpretation. 
Since cifar-100 is prone to overfitting if no augmentation or advanced training protocol is used, it can be seen that the test loss starts to rise quickly after epoch $4$. 
As it is apparent in the plots, NAS also increases almost simultaneously with the test loss: Whilst the NAS in conv1 rises relatively slowly, a rapid jump can already be observed in conv2. 
The jump in NAS is even recognisable shortly before the test loss diverges. 
The observation that conv1 NAS is usually low and that higher NAS can be observed in deeper layers is in line with other work on activation sparsity (i.e. based on counting zeros after ReLU activation)~\cite{10.1109/hpca.2018.00017}.

Since our NAS overfitting estimation is only based on  activation patterns, it does not depend on the satisfaction of the independence hypothesis~\cite{werpachowski2019detecting}.
To illustrate this we created a test loss which violates the independence hypothesis by including $50\%$ dependent images to this set (Figure~\ref{fig:fig4}(a); orange line).
As can be clearly seen, this perturbed test loss does not allow generalisation error estimations, whereas our NAS indicators still diverge accordingly.

To further investigate our observation that NAS coincides with overfitting, we used dropout since this technique is known to reduce generalisation errors.
As shown in Figure~\ref{fig:fig4}(b) dropout indeed delays the divergence of the test loss and also reduces its slope.
However, overfitting cannot be avoided which is also reflected in the median NAS measures throughout the layers. 
Moreover, all but the conv1 NAS measures saturate close to $s_k=1$ after epoch $18$ indicating a drastically reduced number of highly activated features propagating through the network. 
This observation, combined with the constantly increasing test loss value suggests, that the remaining features are over-adapted to the training data.

As theory suggests, applying L1 regularisation encourages sparse weight tensors and should therefore affect our NAS measures. 
As shown in Figure~\ref{fig:fig4}(c), we can see a relatively steep increase of NAS in all layers compared to all other experiments. 
Furthermore, all layers (even conv1) quickly converge towards NAS values close to 1. 
Again, the moment of diverging loss and increasing NAS coincide. 
A similar observation can be made for L2 regularised networks (Figure~\ref{fig:fig4}(d)). 
Even though L2 regularisation is able to shift the moment of overfitting to a later epoch, the corresponding NAS values start to rise slowly before strong generalisation errors are observable in the test loss estimates.

Adding batch normalisation layers to a given network architecture is a widely used approach to prevent overfitting~\cite{Ioffe2015}. 
In line with other work on this regularisation technique the loss does not show a noticeable divergence throughout our training as can be seen in Figure~\ref{fig:fig4}(e).  
Similarly our NAS estimates also tend towards constant values $< 1$ which indicates that batch normalisation has an attenuating effect on activation sparsity over time. 
Note that the first layer (conv1) has no tendency towards sparsified activations throughout the entire training.
We conclude that despite the different effect of batch normalisation on overfitting, the overall tendency of coinciding NAS propagation can also be observed for almost all layers.

\subsection{Spatial Analysis of Activation Sparstiy}
\begin{figure}[t]
   \centering 
   \includegraphics[width=1.0\linewidth]{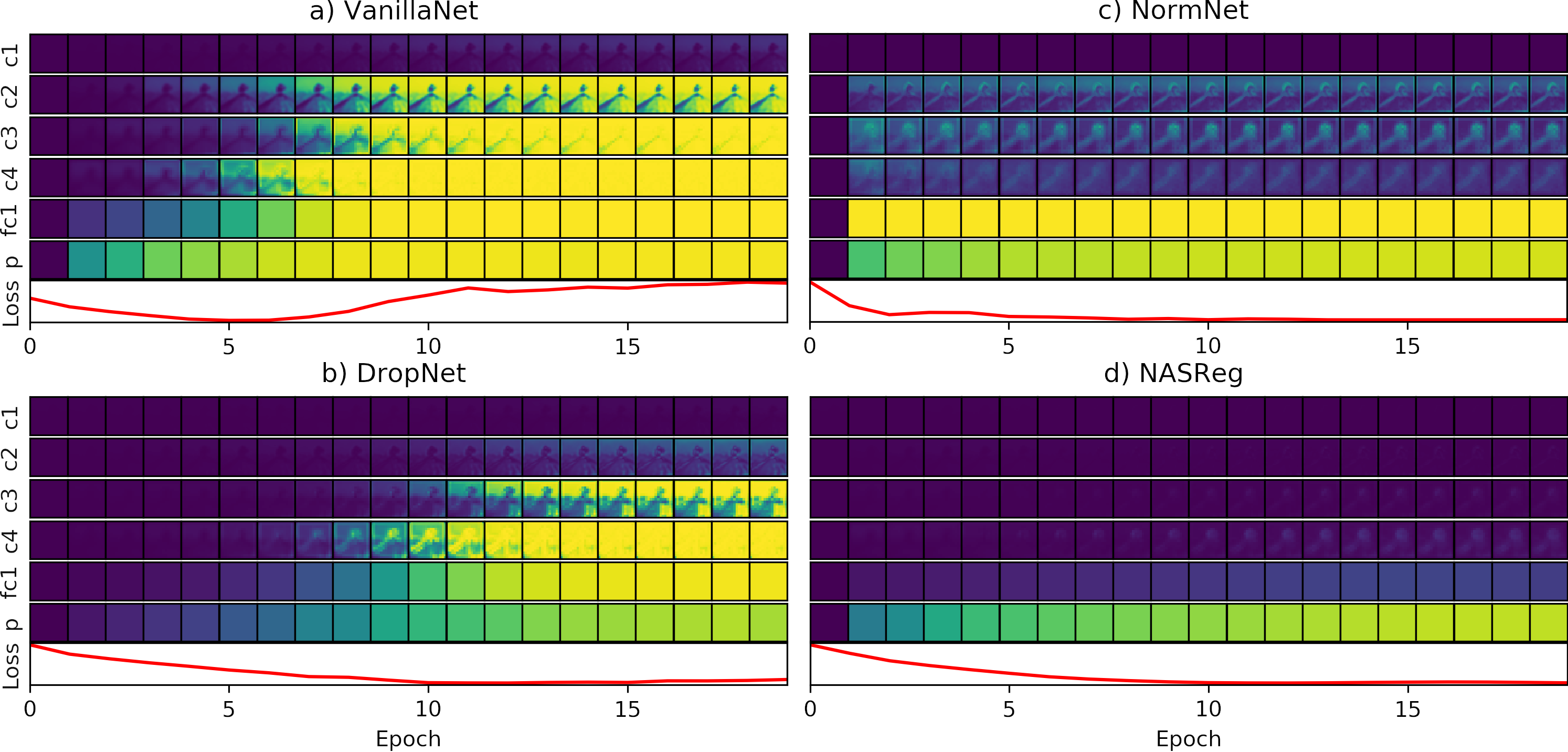}
   \caption{NAS Heatmaps and corresponding test loss values for (a) VanillaNet, (b) DropNet, (c) NormNet and (d)  NASReg  trainings.
	The respective heatmaps were created after each corresponding epoch.
	The original input image can be seen in Figure~\ref{fig:sparsity_visualization} (colour map corresponds to Figure~\ref{fig:vis}). 
}
   \label{fig:fig5}
\end{figure}

Figure~\ref{fig:fig5}(a) illustrates the NAS Heatmaps generated for the individual VanillaNet layers over $30$ epochs of training in comparison to the corresponding test loss.
Similar to the observations made in the previous section the test loss indicates overfitting after epoch $4$ and the conv1 layer is not very susceptible to changes in NAS, whereas the other layers show increased sparsity  proportional to their depth. 
The same phenomenon can be seen in Figure~\ref{fig:fig5}(b) in which we used dropout to delay overfitting. 
In analogy to the previous results, the NAS measures shift accordingly to later epochs but the overall tendency of coinciding generalisation error and increasing NAS is preserved. 
Moreover, spatial NAS representations further reveal the location of sparsified features and therefore less activated filters.

In order to further investigate our spatial NAS visualisation strategy, we visualised these dynamics for a batch normalisation regularised network (Figure~\ref{fig:fig5}(c)).
Batch normalisation can effectively counteract overfitting as can be seen in both, the loss plot and the corresponding NAS heatmaps.
The heatmaps reveal that the overall NAS is low and does not show a significant change over time.
However, subtle features of the original input appear relatively early in deeper conv layers highlighting the strengths of this more detailed visualisation technique.

In summary, these observations suggest that a progressive escalation of NAS tends to coincide with overfitting.
To further investigate both, this hypothesis and the validity of our sparsity definition, we are penalising high NAS values in a layer-wise fashion in the next experiment.

\subsection{Activation Sparsity Regularisation Results}
To study the impact of activation sparsity on overfitting in more detail, we applied NASReg to enforce low NAS values in individual layers except the prediction layer.
Figure~\ref{fig:fig4}(f) shows the course of a NAS regularised training.
Note that in comparison to Figure~\ref{fig:fig4}(a) - (e) we substantially extended training to $300$ epochs to show potential long-term effects. 
Since the regularisation adds a penalty term to the loss, the resultant cross-entropy loss is shown here.
Juxtaposing NASReg with the other trainings, one can clearly see the effect of the regularisation on the  recorded training history. 
VanillaNet, DropNet, L1Net and L2Net clearly start overfitting before the $20$th epoch. 
In comparison, the NASReg test loss measures does not indicate any generalisation errors even when trained for several hundreds of epochs. 
Moreover this training reaches a significantly lower loss value compared to the other trainings. 
One can observe that the NAS regularised layers reveal a different behaviour in comparison to the other trainings: 
Despite the initial increase in NAS estimates the regulariser pushes all NAS measurements towards values close to zero shortly before epoch~$15$.

The differences between NASReg and the other trainings is also shown in Figure~\ref{fig:fig5}(d). 
Here one can see a strong similarity to the NormNet training, which also does not suffer from overfitting in the given epochs. 
Since we explicitly encourage low NAS values using NASReg the heatmaps given in Figure~\ref{fig:fig5}(d) comprise even lower activation sparsity values than NormNet.
Moreover, we regularise fc1 so that only the final prediction layer learns to produce sparse activations, as desired for conventional classification tasks.
Even though we used intentionally straight forward models the positive influence of regularised NAS can also be observed in the test accuracy:
As can be seen in Figure~\ref{fig:fig4} the peak accuracy of NASReg ($0.5119$) outperforms VanillaNet ($0.3851$), DropNet ($0.4034$), L1Net ($0.3267$) and L2Net ($0.2866$).
The smaller accuracy difference to NormNet ($0.4947$) follows the intuition that a certain degree of activation sparsity is acceptable in deeper layers.

\subsection{Activation Sparsity in Common Deep Architectures}

\begin{figure}[t]
     \centering
    \includegraphics[width=1\textwidth]{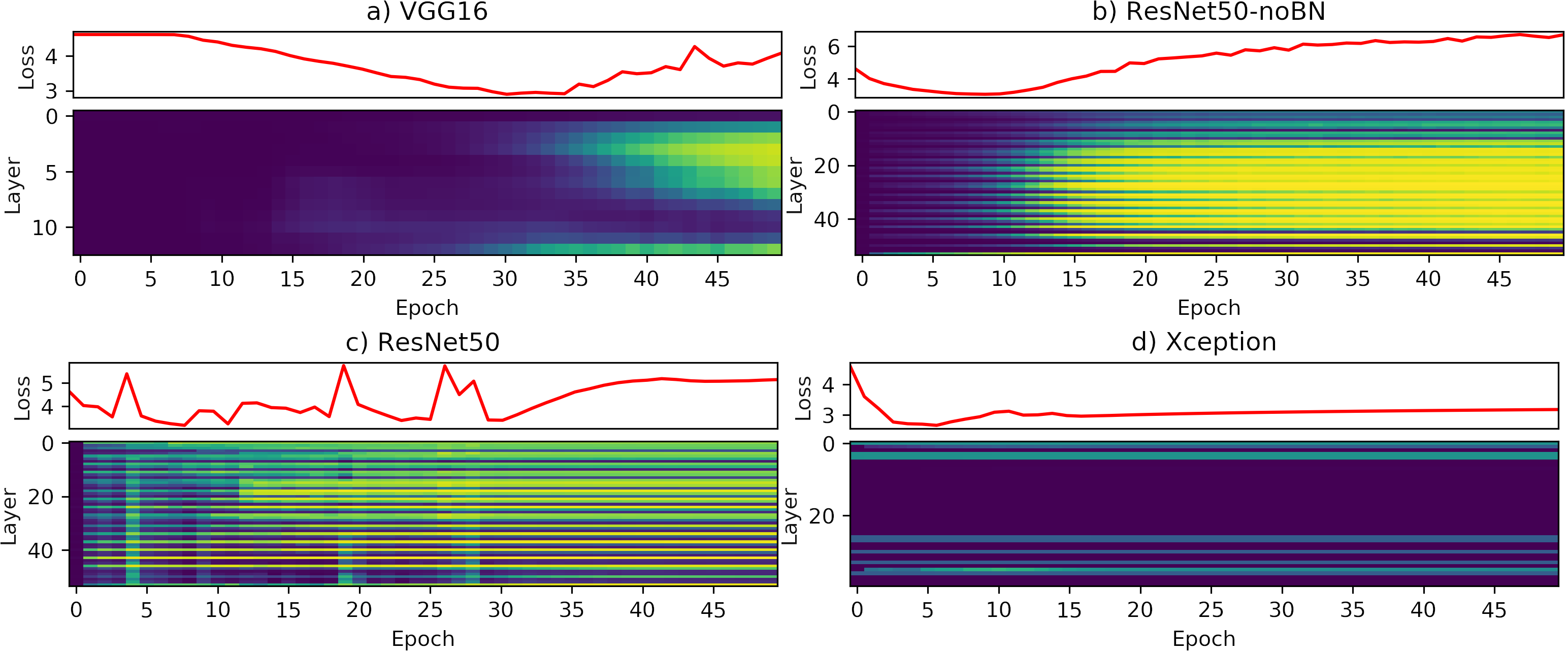}
\caption{NAS Stripe Plots and corresponding test loss (a) VGG16, (b) ResNet50-noBN, (c) ResNet50 and (d) Xception.
Colour map equivalent to Figure~\ref{fig:vis}.
}
\label{fig:results-common}
\end{figure}

After having examined the VanillaNet baseline architecture and its regularised variations, we next investigate further NAS measurements in common deep architectures, namely VGG16, ResNet50 and Xception. 
The training was performed on cifar-100 using the same conditions as described in Section \ref{sec:design}.

The VGG16 network mainly differs from VanillaNet by having a higher number of convolutional layers and a more advanced pooling strategy. 
In line with our hypothesis we can observe the same NAS behaviour when the network starts overfitting (Figure~\ref{fig:results-common}(a)). 
Here, overfitting takes place at around epoch 30 and we can observe increased NAS measures especially in the middle layers shortly before epoch 30.

The ResNet architecture is another common deep CNN, consisting of multiple layers with several shortcut connections. 
Since batch normalisation has a strong influence on activation sparsity (see Section \ref{sec:results-relationship}) we investigated the original regularised ResNet50 and an unregularised version (ResNet50-noBN).
The main difference between ResNet50-noBN and VGG16 is the higher number of layers and the presence of shortcuts. 
Again, as soon as the training drifts into overfitting, the NAS starts rising significantly in most layers (see Figure~\ref{fig:results-common}(b)). 

When applying batch normalisation to ResNet50, we observe an impact on both the NAS and the overfitting measures (see Figure~\ref{fig:results-common}(c)).
Similar to the results presented for the VanillaNet  (cf. Figure~\ref{fig:fig4}), continuous test loss increases can be delayed while achieving better overall loss values when compared to the non-regularised version.
Moreover, batch norm again stabilises many layers to constant NAS values throughout the training (cf. Section~\ref{sec:results-relationship}).
However, the loss rises sharply at epoch 5, 19 and 26, which is also reflected by a sudden NAS increase for most layers.
Overall a more chaotic NAS and test loss behaviour can be observed until the network starts to overfit in epoch 30. 

The overall relationship between overfitting and NAS can also be confirmed in the Xception architecture (see Figure~\ref{fig:results-common}(d)):
When compared to VGG16 and ResNet50 the Xception network achieves best overall generalisation performance, which is also reflected in the low activation sparsity measures.
Since these networks also utilise batch normalisation in most layers we can also observe stabilised NAS values over time.
Interestingly, even in this complex architecture subtle overfitting tendency starting at about epoch 5 can be observed in a deeper layer as slight increases in the NAS estimates are visible at the moment of overfitting which again demonstrates the sensitivity of our XAI measures.


\section{Conclusion}
\label{sec:conclusion}
In this paper we introduced a more sensitive and less biased neural activation sparsity (NAS) definition which we used to derive reliable quantitative measures and layer-wise visualisation techniques to study the dynamics in CNNs during training. 
Interestingly, the analysis of NAS for various network types and regularisation strategies revealed a correlation between activation sparsity and overfitting. 
Throughout our experiments we observed a decreasing amount of discriminative features in almost all layers over time so that the influence of most neurons for the overall classification task became negligible. 
This observation is in line with work on the beneficiary effects of network pruning~\cite{gomez2019learningSparse,liu2018rethinking}.
Potential generalisation error tendencies of heavily sparsified models were however not taken into consideration previously.
Our results suggest that high activation sparsity indeed coincides with overfitting which we were able to detect on a per-layer basis.
Moreover, the rise in NAS values even occurs shortly before the test loss based overfitting measures and our generalisation error estimates do not require the independence between training and test data.

Using our visualisation strategy we were able to show the network performance with respect to the utilised capacity during training.
These visual investigation gave additional evidence that low NAS can be a desirable characteristic for the feature extraction layers.
In contrast, the final classification layer benefits from high activation sparsity to ensure distinct classification decisions.
Applying the NAS measures as a penalty term during training we derived a novel regulariser which prevents CNNs from overfitting over the entire course of training.
In fact, our NAS-based regularisation technique not just supports hypothetical relationship between overfitting and activation sparsity, but also outperforms dropout and batch normalisation in our experiments.

We studied the effects of NAS in a variety of different well-known architectures and regularisation strategies and demonstrated that comparable effects can be observed across all experiments.
These experiments gave additional evidence for the relation between sparsity and overfitting and revealed interesting sparsity-related activation dynamics of batch normalisation.
In the future we will investigate these dynamics in more detail, experiment with more complex network architectures, analyse NAS values for adaptive optimisers such as ADAM and study the potential of our novel regulariser for overall performance improvements.

\label{sec:acknowledgement}
\paragraph{\textbf{Acknowledgements.}} 
BR would like to thank the \textit{Ministeriums für Kultur und Wissenschaft des Landes Nordrhein-Westfalen} for the AI Starter support (ID 005-2010-005). 
Moreover, the authors would like to sincerely thank Sören Klemm for his valuable ideas and input throughout this project and the \textit{WWU IT} for the usage of the \textit{PALMA2} supercomputer.  This work was partially supported by the \textit{Deutsche Forschungsgemeinschaft} (DFG) under contract LI 1530/21-2.  

%
%
%
%
\bibliographystyle{splncs04}
\bibliography{bib}




%
\end{document}